\documentclass[pdflatex,sn-vancouver,Numbered]{sn-jnl}

\usepackage{makecell}

\usepackage{graphicx}%
\usepackage{multirow}%
\usepackage{amsmath,amssymb,amsfonts}%
\usepackage{amsthm}%
\usepackage{mathrsfs}%
\usepackage[title]{appendix}%
\usepackage{xcolor}%
\usepackage{textcomp}%
\usepackage{manyfoot}%
\usepackage{booktabs}%
\usepackage{algorithm}%
\usepackage{algorithmicx}%
\usepackage{algpseudocode}%
\usepackage{listings}%

\begin{document}

\title[Article Title]{\textbf{Bias correction of wind power forecasts with SCADA data and continuous learning}}

\author*[1,2]{\fnm{Stefan} \sur{Jonas}}\email{stefan.jonas@bfh.ch}
\author[3]{\fnm{Kevin} \sur{Winter}}
\author[3]{\fnm{Bernhard} \sur{Brodbeck}}
\author[1, 4]{\fnm{Angela} \sur{Meyer}}

\affil[1]{\orgdiv{School of Engineering and Computer Science}, \orgname{Bern University of Applied Sciences}, \orgaddress{\street{Quellgasse 21}, \city{2501 Biel}, \country{Switzerland}}}

\affil[2]{\orgdiv{Faculty of Informatics}, \orgname{Università della Svizzera italiana}, \orgaddress{\street{Via la Santa 1}, \city{6962 Lugano-Viganello}, \country{Switzerland}}}

\affil[3]{\orgname{WinJi AG}, \orgaddress{\street{Badenerstrasse 808}, \city{8048 Zurich}, \country{Switzerland}}}

\affil[4]{\orgdiv{Department of Geoscience and Remote Sensing}, \orgname{Delft University of Technology}, \orgaddress{\street{Stevinweg 1}, \city{2628 Delft}, \country{Netherlands}}}

\abstract{Wind energy plays a critical role in the transition towards renewable energy
sources. However, the uncertainty and variability of wind can impede its full potential and the
necessary growth of wind power capacity. To mitigate these challenges, wind power forecasting
methods are employed for applications in power management, energy trading, or maintenance
scheduling. In this work, we present, evaluate, and compare four machine learning-based wind
power forecasting models. Our models correct and improve 48-hour forecasts extracted from
a numerical weather prediction (NWP) model. The models are evaluated on datasets from a
wind park comprising 65 wind turbines. The best improvement in forecasting error and mean
bias was achieved by a convolutional neural network, reducing the average NRMSE down to
22\%, coupled with a significant reduction in mean bias, compared to a NRMSE of 35\% from the
strongly biased baseline model using uncorrected NWP forecasts. Our findings further indicate
that changes to neural network architectures play a minor role in affecting the forecasting
performance, and that future research should rather investigate changes in the model pipeline.
Moreover, we introduce a continuous learning strategy, which is shown to achieve the highest
forecasting performance improvements when new data is made available.}

\keywords{wind energy, wind power forecasting, machine learning, bias correction, numerical weather prediction, neural networks}

\maketitle

\section{Introduction}\label{sec1}
Recent growth and progress of renewable energy sources, especially solar and wind, have caused rapid transformative changes in power systems across the world \cite{iea_renewables_2022}. Wind energy in particular, the leading non-hydro renewable energy source, has accounted for a record growth of 17\% in 2021 by 273 TWh, the highest growth amongst all renewables \cite{iea_wind_2022}. However, as wind energy develops into a more critical part of energy grids worldwide, so does the necessity for improved wind power forecasting \cite{haupt_taming_2015}. Generally, the uses of wind speed and power forecasting can be divided into subgroups by the considered forecasting time horizon \cite{jung_current_2014}: Very short-term (up to 30 minutes) forecasts can be employed for turbine control and load tracking, a short-term (up to 6 hours) horizon with applications for preload sharing, medium-term (6 to 24 hours) for power system management and energy trading, and long-term (1-7 days) forecasting used for purposes such as maintenance scheduling, the most relevant to our presented work, in which we will present wind power forecasting models with a forecast horizon of 48 hours.

\subsection{Previous work}
The wind speed or power forecast task can be split into methods based on physical models and statistical models \cite{jung_current_2014}: Physical models, such as numerical weather prediction (NWP) models, provide forecasts for wind speeds and other meteorological variables by computationally solving physical equations describing atmospheric processes. These models are computationally expensive and typically provide a limited resolution. Forecast data derived from NWP models can contain inherent uncertainties and biases, coming from the model formulation, simplification of physics, initial measurements, or surface characteristics \cite{al-yahyai_review_2010}. Statistical methods aim to model the relationship between the target variable (e.g., wind speed or turbine power output) and the available observed data using statistical (learning) techniques. There has been extensive research of pure statistical methods for wind speed and power forecasting, i.e., models which output forecasts based on only the historical wind speeds and other meteorological observations. The proposed techniques range from more traditional statistical techniques such as ARIMA \cite{kavasseri_day-ahead_2009}, to modeling the wind speed and power output using stochastic differential equations \cite{louka_improvements_2008}, or to applying the most recent deep learning advances such as N-BEATS \cite{putz_novel_2021} or temporal convolutional neural networks \cite{gan_temporal_2021}. For a more complete overview of these advances in statistical methods, we refer the reader to the reviews of \cite{jung_current_2014, wang_review_2021}.\\
Another subcategory of statistical approaches, into which our work falls into, are statistical methods which incorporate NWP model forecasts as predictor variables. The purpose of these hybrid models is to correct any biases inherent in the NWP model and to enhance the forecasting capabilities of the provided forecasts. Early works such as \cite{larson_short-term_2006} have shown that incorporating NWP model data as predictor variables for the wind power forecasting task can greatly improve hour-ahead forecasts compared to pure statistical models. The results discussed in \cite{louka_improvements_2008} suggest that corrected NWP data from a model with a moderate resolution using a Kalman filter can provide more reliable forecasts than obtaining computationally expensive high resolution NWP data. Examples of the present bias and characteristic errors found in NWP data are analyzed in detail in the works of \cite{yan_wind_2013, pearre_statistical_2018, wang_multi-step-ahead_2017}. The NWP error examples are shown to be systematically linked to the season (months), the forecasted wind speed value itself, and the wind speed direction. Similarly, \cite{xu_short-term_2015} and \cite{qu_short-term_2013} extract specific NWP error patterns and incorporate these groups of errors into their correction models.\\
A prominent correction approach present in earlier works is applying Kalman filtering to correct the bias, as shown in \cite{louka_improvements_2008, zhao_performance_2012, lima_meteorologicalstatistic_2017}. These approaches are restricted to correcting only one variable at a time and cannot incorporate other NWP variables into the model. A benefit of Gaussian process regression is the ability to incorporate multiple input variables into the regression model, such as applying regression using the NWP wind speed, wind direction, humidity, and temperature in \cite{chen_wind_2014}, or additionally with variables regarding atmospheric stability in \cite{hoolohan_improved_2018}. In more recent methods, deep neural networks as forecasting models are prevalent. A standard artificial neural network structure is presented in \cite{xu_short-term_2015, eseye_short-term_2017, donadio_numerical_2021}, where in the latter a novel look-back parameter is introduced to determine how many hours of historically observed wind speed to include in the input variables. LSTM or RNN-based architectures, which are ideal for sequence-based multi-step forecasts, are very commonly used as basis with changing forecasting pipelines. \cite{felder_wind_2010, lopez_wind_2018, fu_multi-step_2018, wu_datadriven_2019, zhang_short-term_2019, salazar_multivariable_2022} introduced a novel power forecasting architecture inspired by denoising autoencoders, further confirming the benefit of incorporating atmospheric variables as input features. A more advanced deep learning architecture was presented in \cite{zhang_multi-source_2021}. The authors proposed an encoder-decoder based temporal-attention network, which learns to selectively choose data from the multi-source NWP dataset, and the extent of historical data to include in a forecasting step. Across literature, there is an emphasis on feature selection and engineering, the complexity of the present NWP bias, and on finding appropriate forecasting model architectures. 

\subsection{Contribution to research}
Despite these numerous advancements, the insights remain limited: As the comparability across
works is naturally very restricted due to differences in turbines, locations, datasets, and forecast
horizons, the best model approach choice remains unclear. The presented results tend to not
be adequately compared to other benchmarks, rather only to traditional approaches, even if
the contribution introduces more advanced models. We provide such a comparison of four
commonly used and yet significantly different machine learning-based models. Moreover, we
introduce continuous learning to the wind power forecasting task. With continuous learning we
refer to a strategy of periodic updating of the model whenever new data has been made available.
To the best of the author’s knowledge, we are the first to investigate the best strategies to update
a wind power forecasting model to the newest data.\\

The paper is structured as follows. Section \ref{sec:case_study} presents the dataset of our case study and section \ref{sec:models} presents the bias correction models. In section \ref{sec:results}, the bias correction results are shown, compared, and discussed. Finally, section \ref{sec:conclusion} offers a conclusion and possible future research directions.

\section{Case study}\label{sec:case_study}
\subsection{Acquired data}
Our dataset contains measurement data of 65 commercial 2.1 MW wind turbines (WTs) from a wind farm in which all turbines share the same manufacturer and technical specifications, as outlined in Table \ref{tbl:wt_specs}. The Supervisory Control and Data Acquisition (SCADA) data provides for each WT 10-minute averages of the measured turbine power output, wind speed, nacelle and wind direction, and the environment temperature for a continuous timeframe of 2 years. We obtained Numerical Weather Prediction (NWP) data for the identical location as the wind farm and for the same timeframe. This data comprises computed forecasts of the wind speed, wind gust, environment temperature, wind direction, radiance, and precipitation. The forecasts have a lead time for up to 72-hours, at an interval of 15 minutes, and are re-calculated every 6 hours. 

\begin{table}[h]
\begin{tabular}{@{}ll@{}}
\toprule
\textbf{Quantity} & \textbf{Value}                                  \\ \midrule
Rotor diameter    & 114 m                                           \\
Nominal power     & 2.1 MW                                          \\
Type              & Variable-speed horizontal-axis pitch-controlled \\
Deployment        & Onshore                                         \\
Gearbox           & three stage                                     \\ \bottomrule
\end{tabular}
\caption{Technical specifications of the 65 identical turbines from our dataset.}
\label{tbl:wt_specs}
\end{table}

\subsection{Data processing}
\textbf{Feature selection and pre-processing.} The forecasts for wind speed, wind gust, wind direction, and temperature from the NWP forecast dataset were retained. From the SCADA dataset, we extracted only the measured turbine power output as target variable. We adjusted all NWP wind speed values to the hub height of the turbines by applying a log-adjustment: $WS_{NWP\ast}=WS_{NWP}\ast\frac{log\left(h_{HUB}\right)}{log\left(h_{NWP}\right)}$ , where $h_{NWP}$ and $h_{HUB}$ are the forecast height of the wind speed by the NWP model and the turbine hub height respectively. 
For each timestep $t$, the following 10 variables were included as predictors: NWP wind speed, NWP wind gust, NWP temperature, NWP wind direction (sin, cos), hour of $t$ as cyclical variables (sin, cos), $t$ as day of year as cyclical variables (sin, cos), and the forecast lead time. We normalized each variable separately, such that each feature value is in the range of $[0,1]$. The model output, i.e., the normalized power prediction, is then de-normalized prior to the model performance evaluation. \\ 

\noindent \textbf{Data matching.}
The NWP and SCADA data were matched so as to create $K$ samples of 48-hour forecasts for each wind turbine separately, where each forecast sample, starting at the unique time $t_0$, is composed of the selected 10 NWP-based input features and the observed turbine power outputs for times $\{t_0, t_{0+1h} …, t_{0+48h}\}$. A forecast sample $k$ consists of the 10 input features based on the NWP forecast data $\mathcal{X}$ for each timestep, and the corresponding true power outputs $\mathcal{Y}$ (target variable): $\mathcal{X}_k={\{NWP_{t_0}\},\ldots,\{NWP_{t_0+48h}\}}$ , $\mathcal{Y}_k={\{P}_{t_0},\ldots,P_{t_0+48h}\}$, where ${NWP_t}$ are the 10 processed NWP-based input features for that specific timestep, and $P_t$ is the observed turbine power output for time $t$.\\ 

\begin{figure}[h]
    \centering
    \includegraphics[width=1.0\linewidth]{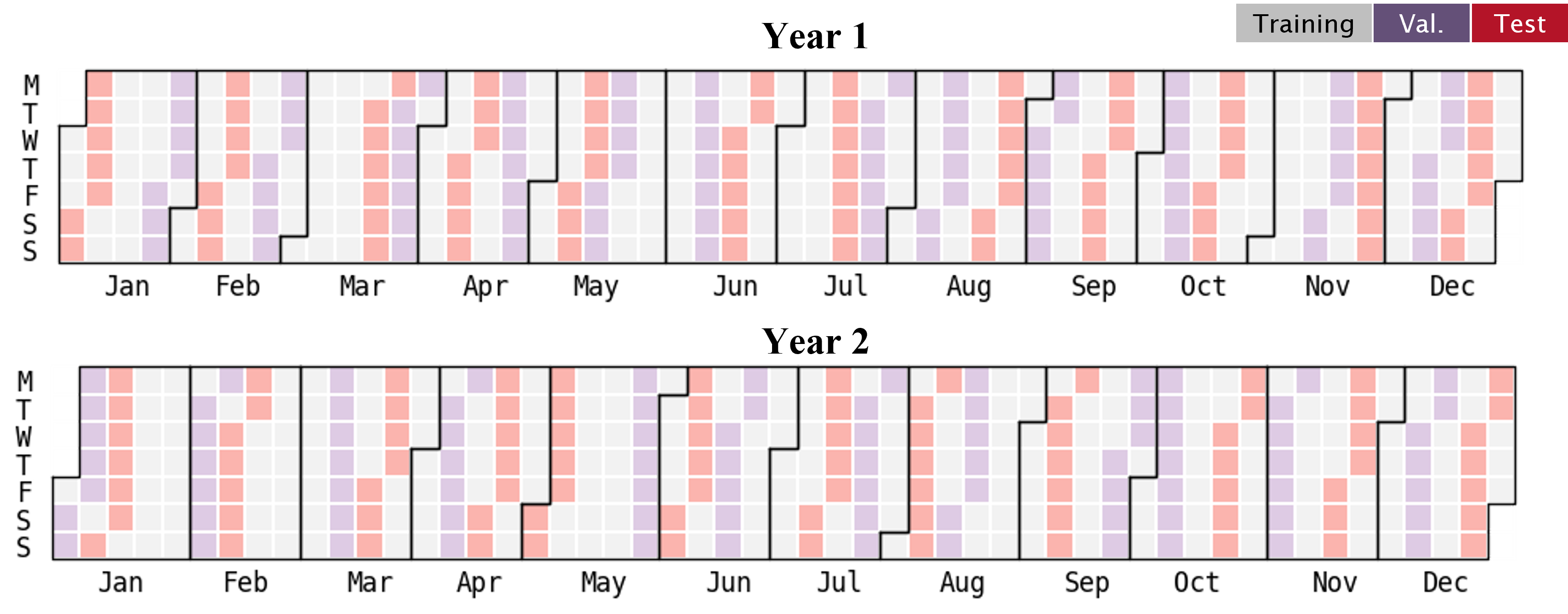}
    \caption{The samples were split according to the days assigned to the training- (grey squared), validation- (purple), and test set (red).}
    \label{fig:dataset_split}
\end{figure}

\noindent \textbf{Dataset split.}
The forecast samples were then split into a non-overlapping training-, validation- and test set. The split was performed monthly, for which randomly chosen 20\% of consecutive days were assigned to the validation set, 20\% of consecutive days to the test set, and the remaining days to the training set. Each turbine’s dataset was split according to the same days assigned to its training, validation, and test set. Figure \ref{fig:dataset_split} illustrates the dataset splits across the two years present in the dataset.

\section{Models}\label{sec:models}
\subsection{Wind power forecasting models}
We introduce four machine learning-based correction models trained to correct the biases and to improve the forecast capabilities compared to an uncorrected baseline model. The models are trained for all 65 WTs in the dataset separately. In general, each of the four presented correction models takes as input the pre-processed NWP data and outputs a wind power forecast. The gradient boosting regression model (GB) and the artificial neural network (NN) output a single wind power forecast for time $t$, based on the NWP features for only the specified single timepoint, while the convolutional neural network (CNN) and the long short-term memory (LSTM) models predict all 48 consecutive hours of wind power data at once, based on the entire 48 hours of NWP features as input sequence in the forecasting sample. Formally, the two single-timepoint models GB and NN are of the form:

$f\left(X_{k,t}\right)=\widehat{Y_{k,t}};\ \ \ \ \ \ k=1,\ldots,K;\ \ \ \ t=t_0,\ldots,t_0+48h$,

\noindent whereas the CNN and LSTM models output a sequence of power forecasts per sample: 

$f\left(X_k\right)=\widehat{Y_k};\ \ \ \ \ \ \ k=1,\ldots,K$\\ 

\begin{figure}[h!]
    \centering
    \includegraphics[width=1\linewidth]{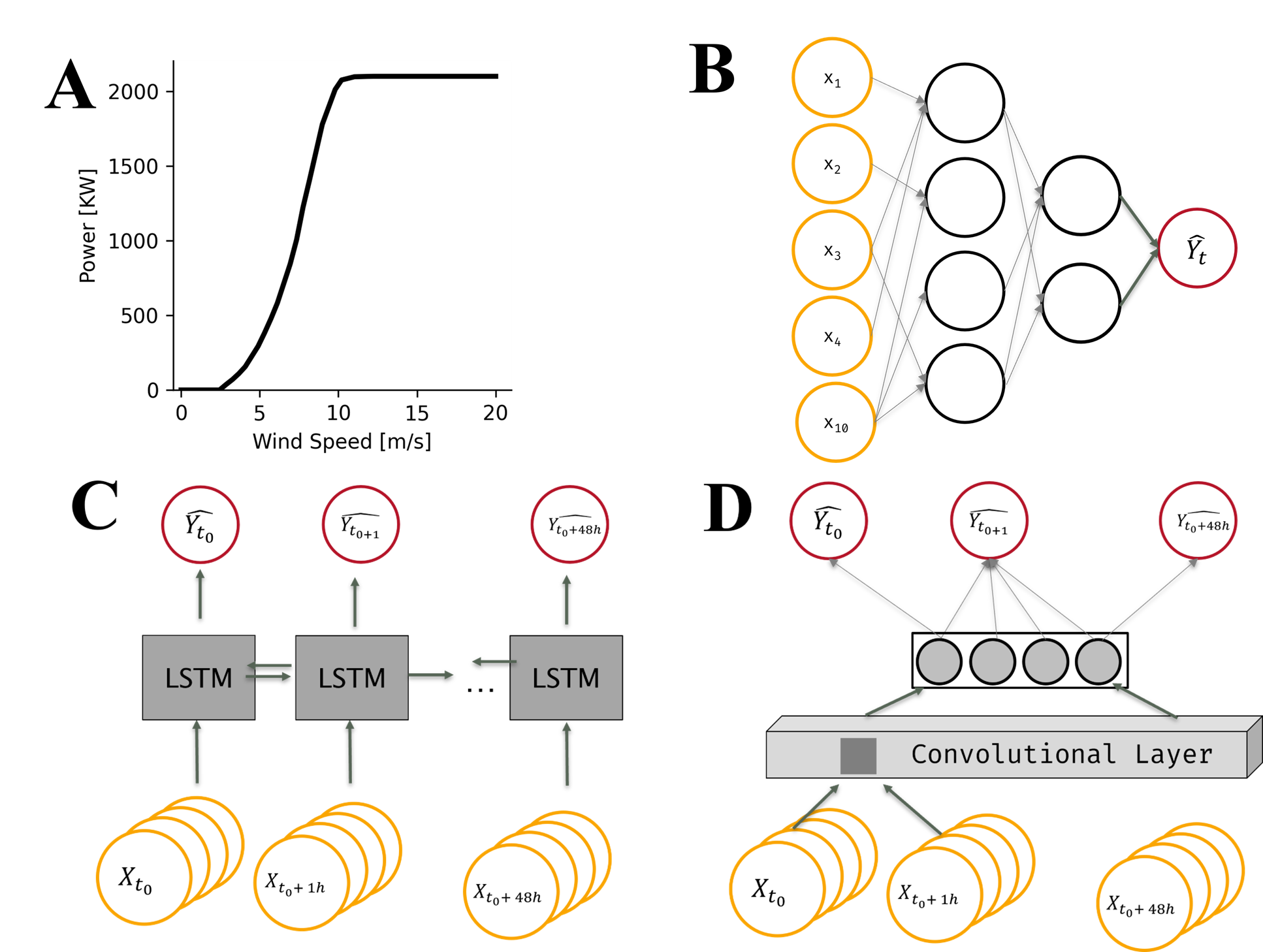}
    \caption{Simplified illustration of four presented models: A: Baseline Model. B: Neural Network. C: LSTM. D: Convolutional Neural Network.}
    \label{fig:correction_models}
\end{figure}

\noindent \textbf{Baseline}. The baseline model represents the uncorrected forecasts based on the NWP data. The baseline model is a power curve provided by the wind turbine manufacturer. The model takes as input the NWP wind speed forecast for a timepoint $t$ and outputs an expected wind power output. As the input is not modified, it transfers all the biases inherent in the NWP model to the power output, i.e., it is an uncorrected forecast. This model is used to assess the extent of the bias in the power prediction task and to evaluate the improvements achieved by the other presented approaches. Figure \ref{fig:correction_models}A shows an illustration of the provided power curve.\\

\noindent \textbf{Gradient boosting} \cite{freund_decision-theoretic_1997, friedman_additive_2000, friedman_greedy_2001} is a popular machine learning-based technique to build prediction models founded on ensembles of weak predictors, typically decision trees, used for regression and classification tasks. In this work, we propose a decision tree-based gradient boosting regression model, as described in Table \ref{tbl:gb_model}. This prediction model outputs a wind power prediction for a timepoint $t$, given the ten corresponding processed input features extracted from NWP data. The model aims to correct the NWP forecasts and to output an unbiased wind power prediction. Learning is performed by minimizing the mean squared error loss between the model output and the observed turbine power over all forecast samples in the training set. We trained a separate model for each turbine, i.e., 65 individual gradient boosting regression models were developed.

\begin{table}[h]
\begin{tabular}{@{}ll@{}}
\toprule
\textbf{Model}        & \textbf{Architecture}                            \\ \midrule
Gradient boosting     & - Input: 10 features (for time $t$)                \\
regression model (GB) & - Output: 1 power output prediction at time $t$   \\
                      & - Number of boosting stages: 100                 \\
                      & - Learning rate: 0.05                            \\
                      & - Maximum depth: 5                               \\
                      & - Loss: mean squared error                       \\
                      & - Other parameters: defaults set by scikit-learn \\ \bottomrule
\end{tabular}
\caption{Model parameters of the gradient boost regression model.}
\label{tbl:gb_model}
\end{table}

\noindent \textbf{Neural network}. The fully-connected neural network (NN) aims to learn to output a corrected wind power forecast given the ten corresponding input features of time $t$. This network consists of three fully-connected layers, as outlined in the full configuration description of Table \ref{tbl:nn_model}. During training, the weights of the neural network are optimized by minimizing the mean squared error loss between the true power outputs and the network outputs. In doing so, the network learns the relationship between the NWP input features and the expected turbine power output. An illustration of the neural network is given in Figure \ref{fig:correction_models}B.\\

\begin{table}[h]
\begin{tabular}{@{}ll@{}}
\toprule
\textbf{Model}    & \textbf{Architecture}                                            \\ \midrule
Artificial neural & - Input: 10 features (for time $t$)                                \\
network (NN)      & - Block 1: Dense layer, 64 units, ReLU, BatchNorm, Dropout (50\%)  \\
                  & - Block 2: Dense layer, 64 units, ReLU, BatchNorm, Dropout (50\%)  \\
                  & - Output Layer: Dense layer, 1 unit, linear activation             \\
                  & - Output: 1 power output prediction for time $t$                     \\
                  &                                                                  \\
                  & - Optimizer: Adam (learning rate: 0.003, loss: mean squared error) \\
                  & - Number of trainable parameters: 5,185                            \\ \bottomrule
\end{tabular}
\caption{Configuration of the artificial neural network (NN). ReLU: Rectified Linear Unit \cite{nair_rectified_2010}. BatchNorm.: Batch Normalization \cite{ioffe_batch_2015}. Dropout (Dropout rate): \cite{srivastava_dropout_2014}. Adam: \cite{kingma_adam_2017}.}
\label{tbl:nn_model}
\end{table}

\noindent \textbf{Long short-term memory network}. This model is based on a long short-term memory network architecture (LSTM) \cite{gers_learning_2000, hochreiter_long_1997}. Specifically, we propose a model using a bidirectional LSTM layer \cite{graves_bidirectional_2005} as described in Table \ref{tbl:lstm}. LSTM architectures are ideally suited for sequences of data. As such, and unlike the previously presented models, the LSTM input is a sequence of input features from the entire 48-hour forecast sample and the model outputs a power prediction for all timepoints in that sequence. Additional to the relationship between the input features and the resulting power for a specific timepoint, the LSTM layer attempts to learn information present within the sequence of data, i.e., between features of different timepoints. A bidirectional layer is capable of extracting information in the forward direction of the sequence (from oldest forecasts to newest) as well as in the backward direction. An illustration of this network is shown in Figure \ref{fig:correction_models}C.

\begin{table}[h]
\begin{tabular}{@{}ll@{}}
\toprule
\textbf{Model}        & \textbf{Architecture}                                            \\ \midrule
Long short-term       & - Input: 49 timesteps x 10 features                                \\
memory network (LSTM) & - Block 1: Bidirectional LSTM layer, 96 units                      \\
                      & - Block 2: Dense layer, 16 units, ReLU, BatchNorm., Dropout (50\%) \\
                      & - Block 3: Dense layer, 32 units, ReLU, BatchNorm., Dropout (50\%) \\
                      & - Output layer: Dense layer, 49 units, linear activation           \\
                      & - Output: 1 power output prediction for each 49 timesteps (49 x 1) \\
                      &                                                                  \\
                      & - Optimizer: Adam (learning rate: 0.001, loss: mean squared error) \\
                      & - Number of trainable parameters: 85,937                           \\ \bottomrule
\end{tabular}
\caption{Full architecture of the long short-term memory network used in this work.}
\label{tbl:lstm}
\end{table}

\noindent \textbf{Convolutional neural network}. Convolutional neural networks (CNNs) are prevalent deep learning architectures with a wide variety of applications, particularly successful in computer vision tasks \cite{goodfellow_deep_2016, lecun_backpropagation_1989, li_survey_2021}. Convolutional layers within the network are capable of learning to extract important low- and high-level features of grid-like input shapes. In this work, we propose a network based on 1-D convolutional layers, where the input to the layer is a 1-D timeseries consisting of the sample timesteps $t_0$ to $t_{0+48h}$ (width of the grid) and the corresponding ten input features of each timepoint (channels).  Similar to the LSTM network, the model aims to learn temporal relationships between input features at different timepoints. The configuration is described in Table \ref{tbl:cnn}, and the model is illustrated in Figure \ref{fig:correction_models}D. 

\begin{table}[h]
\begin{tabular}{@{}ll@{}}
\toprule
\textbf{Model}       & \textbf{Architecture}                                                    \\ \midrule
1-D Convolutional    & - Input: 49 timesteps x 10 features (width x channels)                     \\
neural network (CNN) & - Block 1: 1D Conv. layer, 40 filters, width 5, stride 2, ReLU, BatchNorm. \\
                     & - Block 2: 1D Conv. layer, 64 filters, width 5, stride 2, ReLU, BatchNorm. \\
                     & - Flatten layer                                                            \\
                     & - Block 3: Dense layer, 96 units, ReLU, BatchNorm., Dropout (50\%)         \\
                     & - Output layer: Dense layer, 49 units, linear activation                   \\
                     & - Output: 49 timesteps x 1 power output prediction                         \\
                     &                                                                          \\
                     & - Optimizer: Adam (learning rate: 0.0005, loss: mean squared error)        \\
                     & - Number of trainable parameters: 81,593                                   \\ \bottomrule
\end{tabular}
\caption{The proposed CNN model configuration.}
\label{tbl:cnn}
\end{table}

\newpage

\subsection{Model selection}
The presented model hyperparameters in Tables 2-5 were determined as optimal model configurations through a random search algorithm. We evaluated 200 different model configurations for each of the four model architectures. The configurations with the lowest root mean squared error calculated on the validation set of the same randomly selected wind turbine were chosen as optimal hyperparameters.

\subsection{Training procedure}
All models were trained separately on the training set of each wind turbine. That is, we trained 4 individual correction models for all 65 WTs (260 models). During training, the mean squared error between the model output (power prediction) and the true observed turbine power output was minimized. In all neural network cases, early stopping was implemented through which the training was stopped once the validation set loss had not improved within 15 epochs. 

\subsection{Model implementation}
This work was implemented in Python v3.10. For the models, we used scikit-learn v1.2.2 \cite{pedregosa_scikit-learn_2011} and Keras v2.11 \cite{chollet_keras_2015}. All experiments were run on an Intel Xeon CPU @ 2.3 GHz and an NVIDIA T4 GPU.

\section{Results and discussion}\label{sec:results}
\subsection{Bias analysis}
\begin{figure}[h!]
    \centering
    \includegraphics[width=1\linewidth]{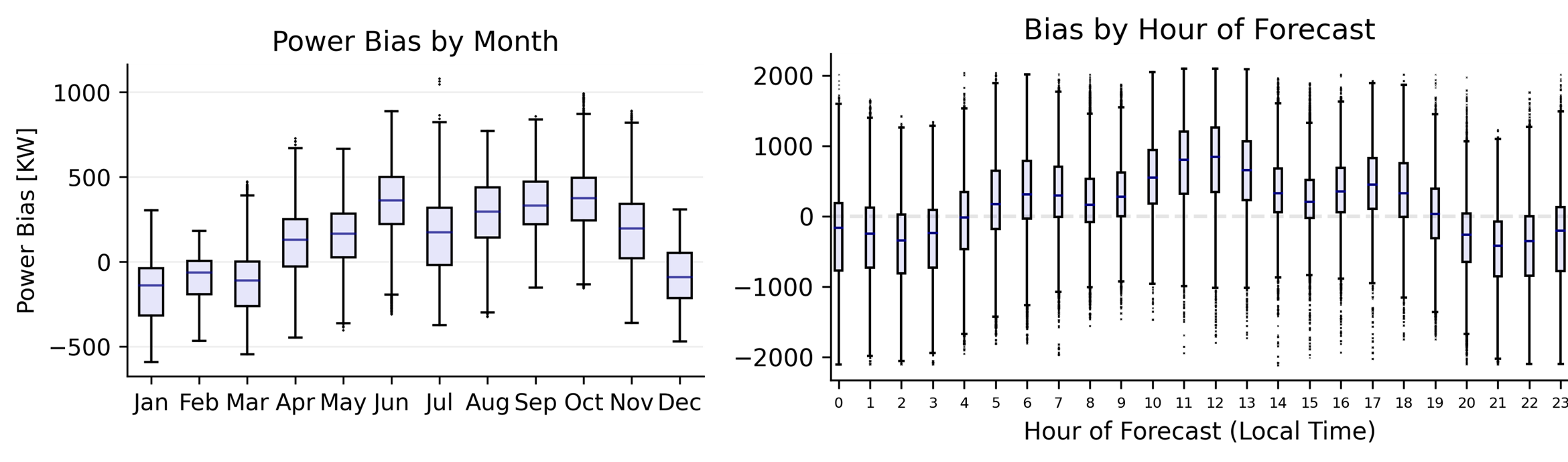}
    \caption{The power bias (forecasted – observed power) calculated with the forecasts obtained from the uncorrected baseline model over the dataset of a single wind turbine. Left side: The power bias depending on the month of the forecast. Right side: The power bias by the hours of the forecast, in the local time zone.}
    \label{fig:biases41}
\end{figure}

\begin{figure}[h!]
    \centering
    \includegraphics[width=.7\linewidth]{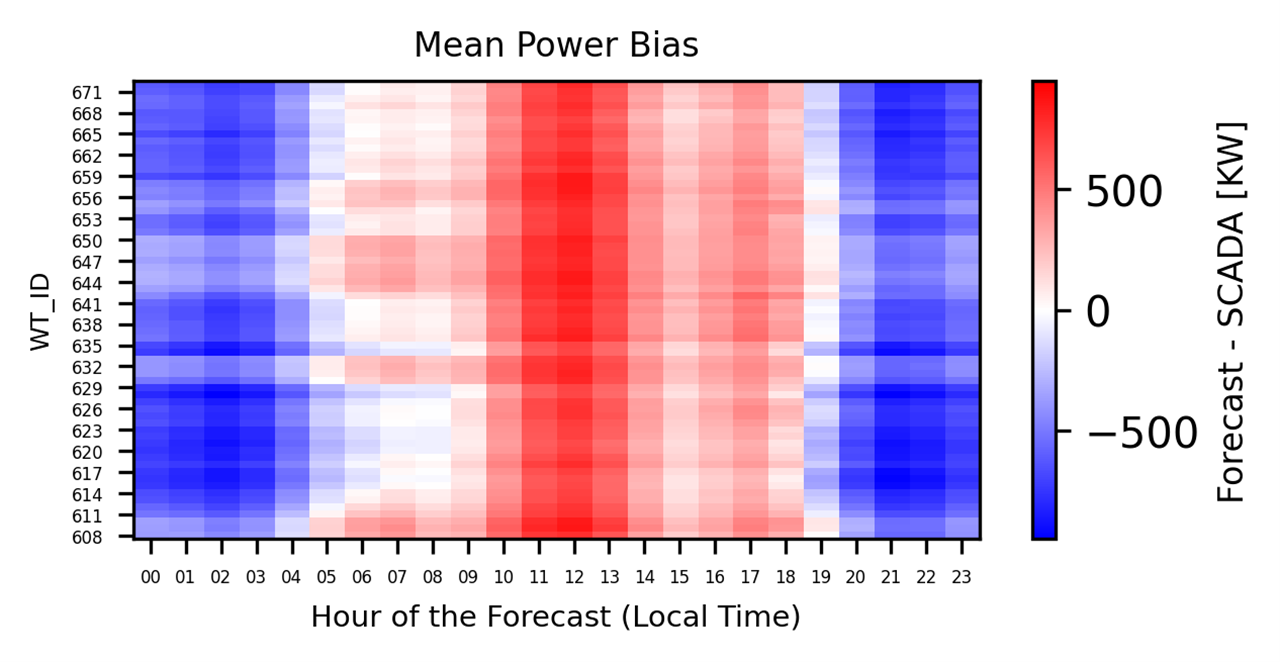}
    \caption{The mean power bias by forecast hour resulting from the forecasts of the uncorrected baseline model. Each row represents the mean bias for one wind turbine. A significant systematic deviation during daytime and nighttime hours is present for all turbines. The extent of the biases differs across the wind park.}
    \label{fig:biases_heatmap}
\end{figure}

Figure \ref{fig:biases41} shows two examples of significant power bias resulting from the power forecasts of the baseline model for one WT. The bias is calculated as the difference between the baseline prediction, i.e., from plugging in the uncorrected NWP wind speed forecasts into the provided power curve, and the true power output from the SCADA data. The figure on the left shows a varying power bias depending on the season of the year, where the power is systematically underestimated (negative mean power bias) by the NWP model in the months December, January, and February, and strongly overestimated in the months June, July, and August. On the right-hand side, an association between the bias and the local hour of the time is observed, where the power is strongly overestimated during daytime hours (6-18) and conversely strongly underestimated during nighttime hours. These two bias cases suggest that the underlying bias drivers are highly complex, as the seasons and times depend on a multitude of features, such as climate, current temperature, the wind speed itself, and many more. Therefore, these results support the inclusion of a multitude of features as input to a correction model, rather than just correcting the wind speed. In Figure \ref{fig:biases_heatmap}, the resulting power bias by local forecast hour across all 65 WTs is presented. While these results confirm that the same previously observed day-night bias patterns are present across all wind turbines, we also find significant individual differences, suggesting the need for individually fitted models rather than one global model. These individual differences can be for instance be due to geospatial differences, individual turbine characteristics (e.g., maintenance history, age), or wake effects.

\newpage 

\subsection{Evaluation metrics}
We trained all proposed models on the training sets and evaluated them on the test sets of each WT. The evaluation data based on $K$ 48h-forecast samples contain the model power predictions  $\widehat{Y_{t_0,k}},\ldots,\widehat{Y_{t_0+48,k}}$, and the corresponding true power outputs $Y_{t_{0_k}},\ldots,Y_{t_0+48,k}$ for each forecast sample $k$. We assess the model performance with the mean bias (MB), the mean absolute error (MAE), and the root mean squared error (RMSE) as follows: \\

$Mean\ Bias=\frac{1}{K}\sum_{K} {MB}_k=\frac{1}{K}\sum_{K}\frac{\sum_{T}\left(\widehat{Y_{t,k}}-Y_{t,k}\right)}{T}$\\

$MAE=\frac{1}{K}\sum_{K}{{MAE}_k}=\frac{1}{K}\sum_{K}\frac{\sum_{T}\left(\widehat{{|Y}_{t,k}}-Y_{t,k}|\right)}{T}$\\

$RMSE=\frac{1}{K}\sum_{K}{{RMSE}_k}=\frac{1}{K}\sum_{K}\sqrt{\frac{\sum_{T}\left(\widehat{Y_{t,k}}-Y_{t,k}\right)^2}{T}}$ \\

$NRMSE=\frac{1}{K}\sum_{K}{{RMSE}_k}\times\frac{100}{C}=\frac{1}{K}\sum_{K}\sqrt{\frac{\sum_{T}\left(\widehat{Y_{t,k}}-Y_{t,k}\right)^2}{T}}\times\frac{100}{C}$ \\

where $\widehat{Y_{t,k}},Y_{t,k}$ represent the model power prediction and the true turbine power output of forecast sample $k$ at timepoint $t$, respectively, and $C$ represents the installed capacity of the wind turbines. 

\subsection{Model evaluations}
\begin{figure}[h!]
    \centering
    \includegraphics[width=1\linewidth]{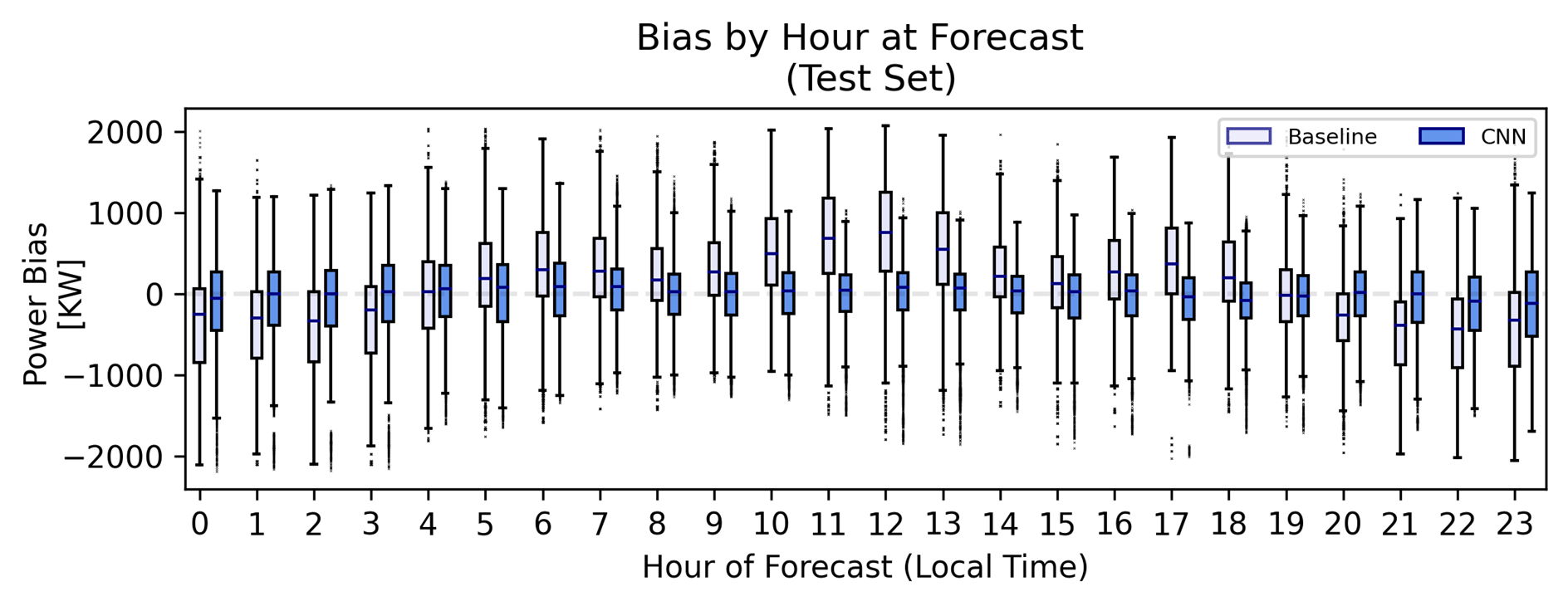}
    \caption{Comparison of the power biases (prediction - observed) by hour of forecast between the uncorrected baseline model and the convolutional neural network model on the test set data of a single wind turbine. The power biases from the uncorrected baseline model (light shaded boxplots on the left) show systematic deviations during day- and nighttime, whereas the power forecasts from the CNN model (dark blue boxplots on the right) show a strong correction of the biases by the model. }
    \label{fig:correction_comparison}
\end{figure}

Our four model architectures were evaluated and compared on the test set of each turbine separately with the specified metrics. In Table \ref{tbl:Results1} the mean bias, MAE, RMSE, and NRMSE obtained on the test for all 65 WTs are listed. The baseline represents the uncorrected forecasting scenario, i.e., the power predictions obtained from plugging in the uncorrected NWP wind speed forecast into the power curve, and it expectedly shows the largest error values. With an average mean bias of -67.2 KW (+- 86), the baseline model shows a very significant bias, as already indicated in the bias analysis of Figure \ref{fig:biases41} and \ref{fig:biases_heatmap}. Additionally, the mean RMSE within a 48h forecast sample with obtained predictions from the baseline model is at a very high value of 725 KW (34.5\% NRMSE), implying unreliable forecasts resulting from the baseline model. Our four machine learning-based correction models all show a significant reduction in bias and errors on the test sets: The gradient boosting regression model (GB) improves the average RMSE by more than 200 KW and shows a reduced but still considerable mean bias of 51 KW (+- 30). This model exhibits the least correction capabilities in all metrics amongst the other three correction models. The neural network architectures (NN, CNN, LSTM) improved the power forecasting significantly better: The lowest bias was achieved by the convolutional network, at only 12 KW (+- 27). Additionally, it significantly improved the forecasts down to an average RMSE of 463 KW (NRMSE of 22\%). In Figure \ref{fig:correction_comparison}, an example of the bias correction by the CNN is shown. Compared to the baseline, which shows a strong systematic deviation based on day- or night-time, the CNN forecasts are corrected, with mean biases close to 0 (unbiased) independent of the forecast hour.  

\newpage 

\begin{table}[h!]
\begin{tabular}{@{}lllll@{}}
\midrule
\textbf{Model} & \textbf{\makecell{Avg. Mean Bias \\ {[}KW{]}}} & \textbf{\makecell{Avg. MAE \\ {[}KW{]}}} & \textbf{\makecell{Avg. RMSE \\ {[}KW{]}}} & \textbf{\makecell{Avg. NRMSE \\ {[}\%{]}}} \\ \midrule
Baseline & -67.2 \footnotesize{{[}+- 85.8{]}} & 585.5 \footnotesize{{[}+- 23.4{]}} & 725.1  \footnotesize{{[}+- 27.3{]}} & 34.5  \footnotesize{{[}+- 1.3{]}} \\

GB & 50.5 \footnotesize{{[}+- 30.4{]}} & 435.8 \footnotesize{{[}+- 15.1{]}} & 519.1  \footnotesize{{[}+- 17.6{]}} & 24.7  \footnotesize{{[}+- 0.8{]}} \\

NN & 39.8 \footnotesize{{[}+- 22.2{]}} & 393.2 \footnotesize{{[}+- 13.4{]}} & 476.9  \footnotesize{{[}+- 16.9{]}} & 22.7  \footnotesize{{[}+- 0.8{]}} \\

CNN & \textbf{11.7} \footnotesize{{[}+-   26.5{]}} & \textbf{375.3} \footnotesize{{[}+- 14.9{]}} & 462.9  \footnotesize{{[}+- 16.8{]}} & \textbf{22.0}  \footnotesize{{[}+- 0.8{]}} \\

LSTM & 44.3 \footnotesize{{[}+- 32.9{]}} & 385.6 \footnotesize{{[}+- 19.0{]}} & \textbf{462.3}  \footnotesize{{[}+- 19.8{]}} & \textbf{22.0}  \footnotesize{{[}+- 0.8{]}} \\ \bottomrule
\end{tabular}
\caption{Test set results of the four proposed correction models compared to the uncorrected baseline. The table shows the average value calculated over the test set for each turbine with the standard deviation in brackets. Baseline: uncorrected power curve model. GB: gradient boosting regression model. NN: fully-connected neural network. CNN: convolutional neural network. LSTM: long short-term memory network.}
\label{tbl:Results1}
\end{table}

\subsection{Model comparisons}
\begin{figure}[h!]
    \centering
    \includegraphics[width=0.75\linewidth]{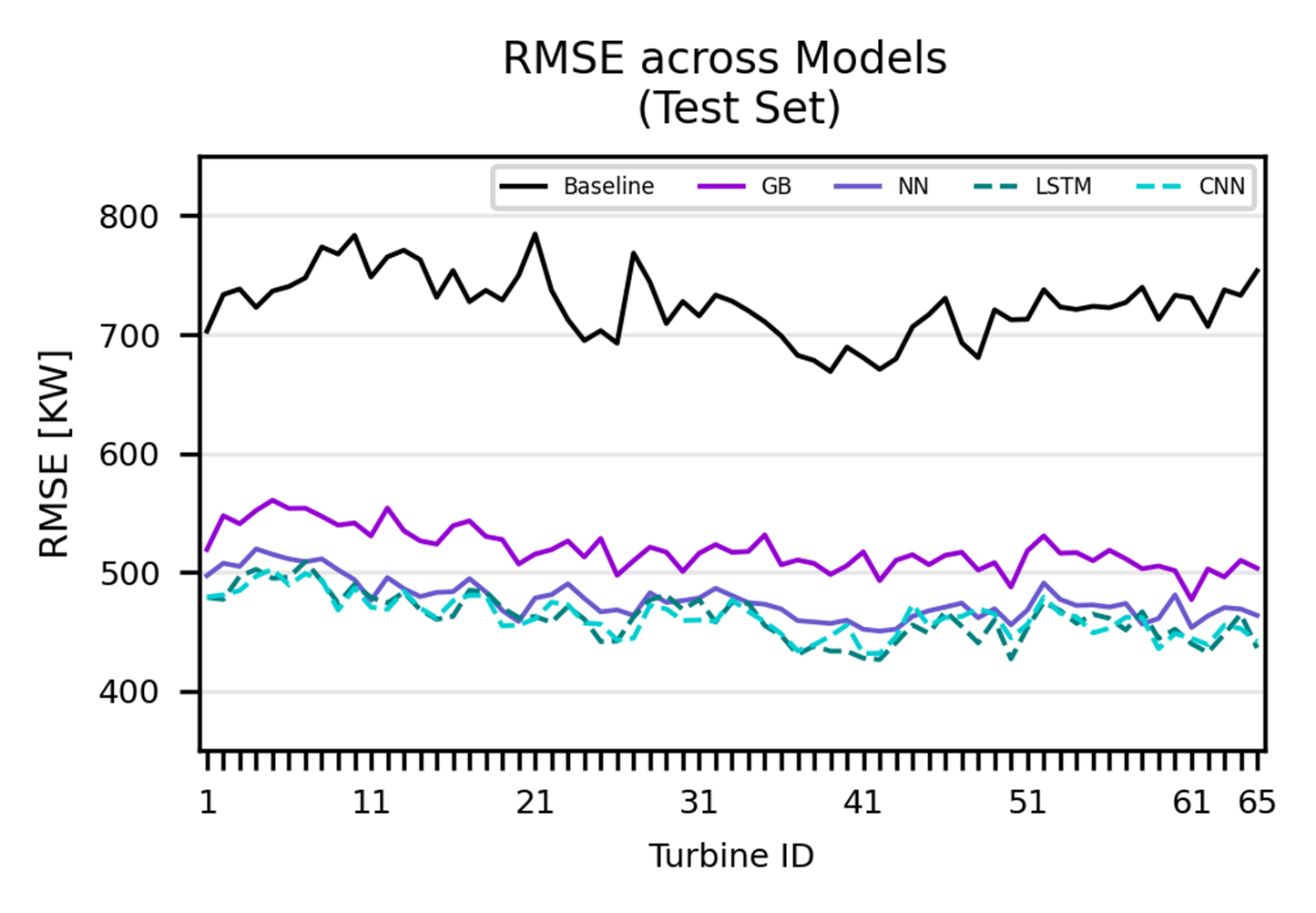}
    \caption{The RMSE calculated over the test set of all 65 wind turbines. For each wind turbine on the x-axis, the RMSE values of the uncorrected baseline model, the gradient boosting regression model (GB), the fully-connected neural network (NN), the long short-term memory network (LSTM), and the convolutional neural network (CNN) are shown. }
    \label{fig:rmses}
\end{figure}

However, there is a limitation in averaging and comparing the absolute error values achieved by the correction models from different WTs, as the forecast quality of the NWP wind speed forecasts can significantly differ across each WT. In Figure \ref{fig:rmses}, the results of the baseline and the correction models are shown for each turbine. While the NWP wind speed forecasts and the power curve are equal for all, they result in notably different forecast errors based on the same power curve, with a wide range from the lowest value of 699 KW (33\% NRMSE) to a maximum of 785 KW (37\% NRMSE). This discrepancy can be attributed for instance to individual turbine differences (e.g., maintenance history, age), terrain characteristics, or wake effects. Thus, the absolute errors from the correction models reflect these discrepancies as well and cannot be perfectly compared. Instead, we compare the forecasting error reduction in relative terms of the baseline, as shown in Table \ref{tbl:inbaseline}. These results first confirm the previous results, showing the least improvement by the gradient boosting regression model at an average 71.6\% (28.4\% error reduction) of the baseline RMSE, and the neural networks performing better and closely to each other, achieving 63.8 – 65.8\% RMSE (an improvement of 34.2 – 36.2\%) of the baseline. Furthermore, the independently trained models across all 65 WTs perform similarly and consistently, as implied by the minor standard deviation of less than 3\% for all models. \\

We note that these four model architectures are considerably different to each other. The gradient boosting model has shown to result in considerably higher forecasting errors compared to all neural network models, and with an average rank of 4.0 out of 4, our results further advocate for the already prevalent use of neural network-based methods for the power forecasting task (Section 1.2). However, we notice that the three different neural networks produce highly similar results. The fully-connected neural network architecture, trained on single timepoints and consisting of only 5,185 trainable parameters, produces only slightly higher RMSE errors (2\% higher relative to the baseline) compared to the other neural networks. The CNN and LSTM models, on the other hand, are trained on complete 48h forecast sequences and contain 81,593 and 85,937 trainable parameters, respectively. Consequently, the benefit of learning the relationship of features across timepoints was only minimal in our case study. Furthermore, regardless of the completely different architectures, these two models perform almost identically in terms of the forecasting ability, as shown by the error values and rankings in Table \ref{tbl:inbaseline}. Our results hint at a limit of the possible bias correction and forecasting improvements with the presented pipeline of correcting NWP-based forecasts. Instead of investigating novel model architectures, future research directions should rather include conceptual changes, as has been already successfully shown by including previous measurements in the input features (e.g., \cite{chen_wind_2014, donadio_numerical_2021, zhang_multi-source_2021}) or by incorporating information from the power curve (e.g., \cite{chen_wind_2014, donadio_numerical_2021}).

\begin{table}[h]
\begin{tabular}{@{}llll@{}}
\midrule
\textbf{Model} & \textbf{\makecell{Avg. MAE in \\ \% of Baseline}} & \textbf{\makecell{Avg. RMSE \\ in \% of Baseline}} & \textbf{\makecell{Avg. Rank \\ by lowest RMSE}} \\ \midrule
GB   & 74.5 \footnotesize{{[}+- 2.8{]}} & 71.6 \footnotesize{{[}+- 2.6{]}} & 4.0 \footnotesize{{[}+- 0.0{]}} \\
NN   & 67.2 \footnotesize{{[}+- 2.5{]}} & 65.8 \footnotesize{{[}+- 2.3{]}} & 2.8 \footnotesize{{[}+- 0.4{]}} \\
CNN  & \textbf{64.2} \footnotesize{{[}+- 2.7{]}} & 63.9 \footnotesize{{[}+- 2.3{]}} & 1.6 \footnotesize{{[}+- 0.6{]}} \\
LSTM & 65.9 \footnotesize{{[}+- 2.6{]}} & \textbf{63.8} \footnotesize{{[}+- 2.3{]}} & \textbf{1.5} \footnotesize{{[}+- 0.7{]}} \\ \bottomrule
\end{tabular}
\caption{Test set results of the four proposed correction models relative to the uncorrected baseline value of each WT. The table shows the average value calculated over the test set for each turbine with the standard deviation in brackets. Baseline: uncorrected power curve model. GB: gradient boosting regression model. NN: fully-connected neural network. CNN: convolutional neural network. LSTM: long short-term memory network.}
\label{tbl:inbaseline}
\end{table}

\subsection{Continuous Learning}
As the datasets in our case study consist of forecasts and observations collected during 2 years, they may lack examples for certain meteorological events (e.g., heatwaves) which did not occur in the available timeframe. The trained correction models on these data may therefore generalize poorly when predicting based on data originating from previously unseen circumstances. There is a clear benefit to update the models to any newly made available data in order to introduce a model to more examples. We introduce the concept of continuous learning (also continual- or lifelong learning) \cite{chen2018continual}, as a strategy to continuously update correction models. Specifically, we propose a procedure similar to the concepts of freezing and fine-tuning weights in transfer learning tasks (e.g., \cite{iman_review_2023}), in which the original weights of the neural networks are used as the initial state, following a slight adjustment (“fine-tuning”) of a subset of layer weights (non-“frozen” layers). With continuous learning, when given new training data, a subset of layer weights is trained on exclusively new data with a smaller learning rate, in order to update the models to new examples, while retaining the learned relationships from the original dataset. As the presented continuous learning strategy relies on neural network weights, it is not applicable to the gradient boosting model. 
We assess the effectiveness of our continuous learning strategy. New data comprising 6 months of forecasts and observations for all wind turbines were collected. The datasets were processed according to the same procedure as the original datasets. We evaluated the forecasting performance on the new test set of one randomly selected WT for the following three strategies: \\
\newline 
\textbf{Strategy 1 – original models}: The models remain unchanged, i.e., we use the neural networks (NN, LSTM, CNN) trained on only the original two years of data to forecast the power output on the test set of the new timeframe. This strategy represents the baseline, where models do not change after new data is made available. \\
\newline
\textbf{Strategy 2 – new models}: We train the three neural networks from scratch exclusively on the new dataset, that is, only on the training set of the new 6 months of our updated dataset. As the training and test set originate from the same timeframe, both sets should have a significant overlap in underlying weather and turbine conditions.  \\
\newline
\textbf{Strategy 3 – continuous learning}: This strategy represents our continuous learning approach. Using the original neural network weights as initial state, we fine-tune selected layer weights by training with the new dataset using a smaller learning rate. \newline

\begin{figure}[h]
    \centering
    \includegraphics[width=0.75\linewidth]{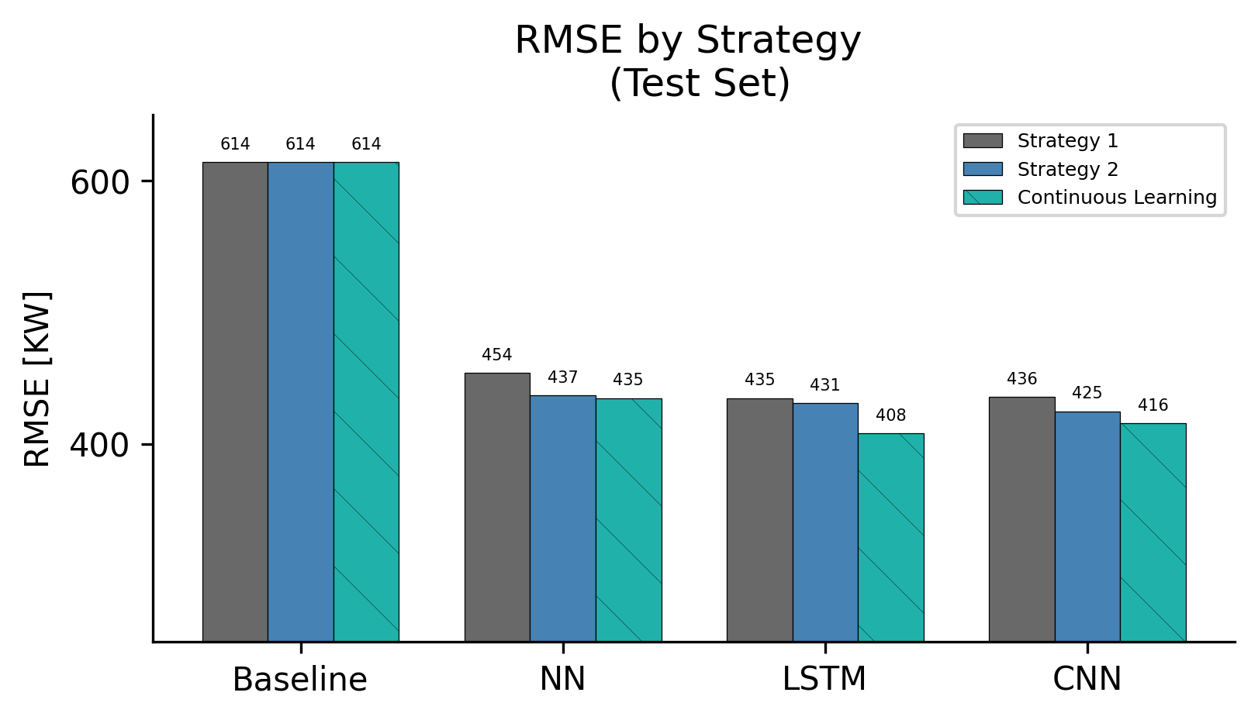}
    \caption{Comparison of the strategy performances on the updated test set of a randomly selected WT.}
    \label{fig:contlearning}
\end{figure}

The results are shown in Figure \ref{fig:contlearning}. We show the RMSE obtained on the new test set of the selected WT. As the uncorrected baseline model only relies on the provided power curve, it is independent of the training data and results in the same performance across all strategies. For the three neural network architectures, we observe the same pattern across each model: Strategy 1 results in the highest RMSE, while in comparison strategy 2 shows a slight reduction in measured error. Our continuous learning approach reaches the best performance in all three cases. As the first strategy uses the original models, these results indicate that there may be some patterns present in the new data which did not occur in the original timeframe, leading to a higher forecast error compared to the other strategies. The improvements in performance achieved by the second strategy show the importance of having the same underlying conditions present in the training data: Despite being trained on significantly less data, the models trained with this strategy show a superior performance to the original models. With our continuous learning strategy, we adjust the pre-trained model weights of the neural networks to the new data, resulting in the models adjusting the previously learned feature relationships to new training instances. Thus, the strategy combines learnt model insights from the original and the new data. \newpage 
Our findings support the use of a continuous learning strategy, with which the models should be iteratively updated whenever new data becomes available, in order to improve the forecast and correction abilities of the models. However, future studies are required to confirm these results and to investigate potential downsides of continuous learning such as catastrophic forgetting \cite{chen2018continual}.

\section{Conclusion}\label{sec:conclusion}
In this work we investigated the task of wind power forecasting based on data extracted from a numerical weather prediction model. The baseline model using uncorrected NWP data was observed to output a strongly biased wind power forecast. Our analysis revealed a strong systematic over- and underestimation based on the seasonality and the time of day, suggesting the necessity of a correction model. We proposed four machine learning-based models to correct biases, namely a gradient boosting regression model, a fully-connected neural network, a long short-term memory network, and a convolutional neural network. All four correction models managed to significantly improve the baseline performance. The best forecasting performance was achieved by the convolutional neural network, although the very close proximity of performance between the neural networks, despite being made up of completely different model configurations, indicates a limit of the possibility of further improving wind power forecasts with the presented forecasting pipeline. Our experiments indicate that future research directions should investigate changes to the model pipeline, such as including different features (e.g., historical SCADA data) or more data sources. Finally, we introduced continuous learning, a strategy to continuously update models when new data becomes available. Our continuous learning strategy proved to achieve the best performance when updated and tested on new data. Further studies with larger datasets are required to assess the benefits and limits of continuous learning.

\bmhead{Acknowledgements}
We gratefully acknowledge the Swiss Innovation Agency InnoSuisse for funding this project.

\bibliography{sn-bibliography}

\end{document}